\documentclass[sigconf]{acmart}

\usepackage{indentfirst}

\usepackage{algpseudocode}
\usepackage{algorithm,algorithmicx}

\usepackage{amsmath}

\usepackage{amssymb}

\usepackage{graphicx}
\usepackage{epstopdf}
\usepackage{stfloats}

\newcommand{\tabincell}[2]{\begin{tabular}{@{}#1@{}}#2\end{tabular}}
\usepackage{multirow}

\AtBeginDocument{%
  \providecommand\BibTeX{{%
    \normalfont B\kern-0.5em{\scshape i\kern-0.25em b}\kern-0.8em\TeX}}}

\copyrightyear{2022}
\acmYear{2022}
\setcopyright{acmcopyright}\acmConference[DAC '22]{Proceedings of the 59th ACM/IEEE Design Automation Conference (DAC)}{July 10--14, 2022}{San Francisco, CA, USA}
\acmBooktitle{Proceedings of the 59th ACM/IEEE Design Automation Conference (DAC) (DAC '22), July 10--14, 2022, San Francisco, CA, USA}
\acmPrice{15.00}
\acmDOI{10.1145/3489517.3530600}
\acmISBN{978-1-4503-9142-9/22/07}

\begin{document}
\title{Worst-Case Dynamic Power Distribution Network Noise Prediction Using Convolutional Neural Network}

\author{Xiao Dong$^1$, Yufei Chen$^1$, Xunzhao Yin$^1$, Cheng Zhuo$^{1*}$, \\$^1$College of Information Science \& Electronic Engineering, Zhejiang University, Hangzhou, China\\$^*$Corresponding Email: czhuo@zju.edu.cn}

\pagestyle{plain}

\begin{abstract}
\noindent 
Worst-case dynamic PDN noise analysis is an essential step in PDN sign-off to ensure the performance and reliability of chips. However, with the growing PDN size and increasing scenarios to be validated, it becomes very time- and resource-consuming to conduct full-stack PDN simulation to check the worst-case noise for different test vectors. Recently, various works have proposed machine learning based methods for supply noise prediction, many of which still suffer from large training overhead, inefficiency, or non-scalability. Thus, this paper proposed an efficient and scalable framework for the worst-case dynamic PDN noise prediction. The framework first reduces the spatial and temporal redundancy in the PDN and input current vector, and then employs efficient feature extraction as well as a novel convolutional neural network architecture to predict the worst-case dynamic PDN noise.  Experimental results show that the proposed framework consistently outperforms the commercial tool and the state-of-the-art machine learning method with only 0.63-1.02\% mean relative error and 25-69$\times$ speedup.

\end{abstract}

\maketitle

\section{Introduction}
\label{sec:introduction}

With the continuous voltage scaling and ever growing integration density, power supply noise has become a concerning issue in modern lower power SoC designs~\cite{2010MissingLink, ZhuoTVLSI2015, ArabiDT2007, ZhuoTCAD2020}. A too large supply noise of the power distribution network (PDN) for an SoC not only degrades the performance of critical circuits, but also impairs its reliability~\cite{2010MissingLink}. Thus, the worst-case supply noise validation has become an essential step in power delivery sign-off~\cite{ZhuoTVLSI2015}. 

\begin{figure}[t]
  \centering
  \includegraphics[width=0.7\linewidth]{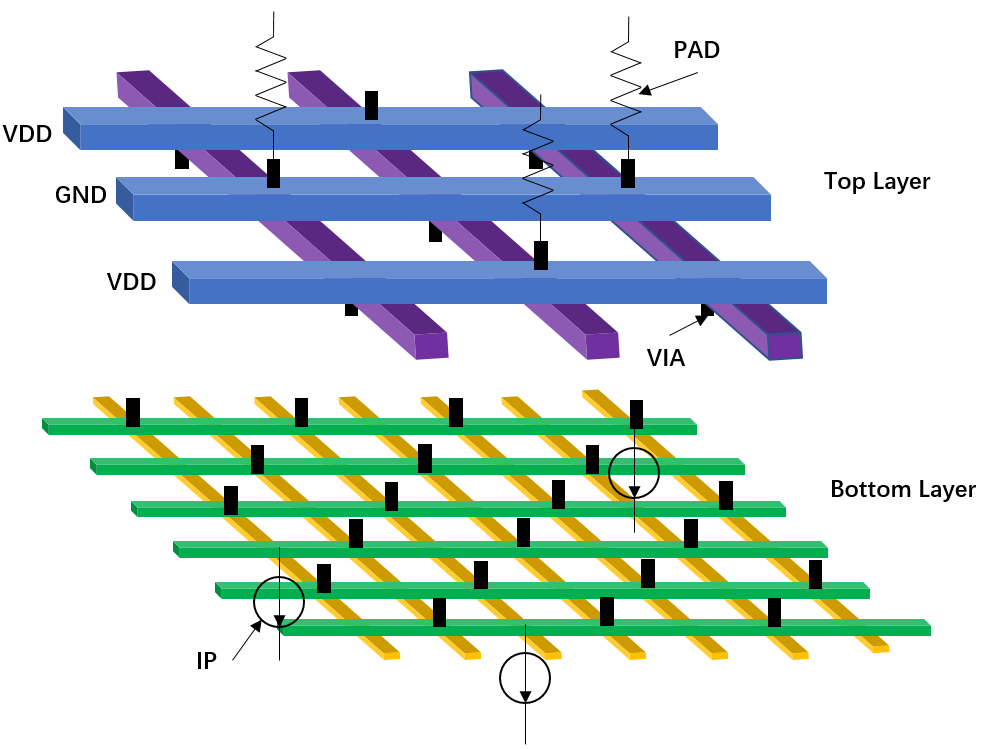}\vspace{-0.4cm}
  \caption{An example of on-die power grid and current excitation.}\vspace{-0.6cm}
  \label{fig:pdn}
\end{figure}

A modern VLSI PDN consists of board, package, and on-die power grid. Fig.~\ref{fig:pdn} shows an illustrative plot of on-die power grid, which may contain more than ten metal layers with wire pitches up to several nano-meter for advanced technologies~\cite{ZhuoTVLSI2015, 2010MissingLink}. Since the worst-case noise validation (WNV) needs to be accurate for sign-off, in commercial practices, WNV is commonly conducted with very detailed distributed RC/RLC model representing the on-die grid along with package and board modelled as macro-models~\cite{ZhuoTVLSI2015}. Obviously, such a simulation needs to solve the voltages for millions to billions of nodes in the on-die grid, which is a non-trivial problem. It has been reported that, for 32nm commercial memory controller, even the commercial tool takes up to 24 hours to solve a few hundred nano-second trace when accounting for the entire PDN from board to die~\cite{ZhuoTVLSI2015}. On the other hand, since there are many application scenarios for an SoC, WNV needs to be executed for tens of test vectors during sign-off, which \textit{makes it very time- and resource-consuming}~\cite{ArabiDT2007}.

Mathematically speaking, given a test vector (e.g., in the form of switching current), WNV is to solve a sparse linear system (representing the entire PDN) and find the worst case noise both temporally (along the vector) and spatially (across the entire power grid). In the past two decades, many works have been proposed to speed up the sparse linear system solve, from multigrid, random walk, to graph spectral sparsification~\cite{2000HA,2003AAMA,2006RandomWalks, 2008AlgebraicMultigrid, ZhaoICCAD2014}.

Recently, with the popularity of machine learning, a few works deployed various machine learning techniques to accelerate the computation of supply noise or electro-migration prediction without actually invoking the sparse linear system simulator~\cite{2020XGBIR,  2021Thermal, 2019IncPIRD, 2020PowerNet, 2018ECO, 2018MLECO,  ZhouICCAD2020, JinDAC2021}. In order to achieve accurate prediction, the detailed PDN structural and electrical information are extracted as input features to train the model, such as instance power and path resistance~\cite{2020PowerNet, 2019IncPIRD, 2018ECO, 2018MLECO}. However, such instance-level information and path resistance itself demands power or static power delivery analysis, which actually involves implicit but \textit{non-trivial training overhead.} In addition, while most existing works placed focus on static IR drop of PDN~\cite{2019IncPIRD,2020XGBIR,2021Thermal}, \textit{dynamic (or transient) PDN noise} is actually more important for sign-off, which is triggered by the resonance between package and die and hence results in more severe noise that needs to be validated. Finally, since the size of a commercial PDN can be tremendous (millions to billions of nodes), the direct employment of deep learning techniques to predict node voltage may easily result in a too huge neural network with serious \textit{scalability} issue~\cite{2021Thermal}. Thus, many works have to either simplify the input to the neural network or check the noise region by region~\cite{2020PowerNet,2018ECO,2018MLECO}, inevitably compromising accuracy or efficiency. Thus, even with many prior machine learning works, it still \textbf{remains an open question to design an efficient and scalable framework to predict the worst-case dynamic noise for the entire PDN without incurring too much training overhead.}

In this paper, we address the above issues and propose an efficient, scalable yet accurate framework to predict the worst-case dynamic PDN noise. The framework accounts for the redundancy both temporally and spatially. It then employs much simpler feature extraction strategy to reduce the training overhead. Finally, a novel convolutional neural network (CNN) architecture is designed to accurately incorporate all the information and predict the worst-case PDN noise. The contributions of our work are summarized as below:
\begin{itemize}
\item We design a scalable and efficient framework for dynamic worst-case PDN noise analysis. It consists of three subnets and can be trained with a small set of randomly produced test vectors. The trained model can support fast dynamic noise analysis with the given test vector and hence speeds up the repeated validations of worst-case PDN sign-off.
\item In order to accelerate the prediction, an algorithm for current feature compression is presented. By filtering out the irrelevant segments in the current sequences, the efficiency can be improved without sacrificing much accuracy.
\item We propose to select the load current and distance to power bumps as the input features, which are easily accessible and provide sufficient information for accurate simulation. 
\item Our framework incorporates a novel CNN architecture to predict the dynamic worst-case PDN noise. The framework can predict the noise map of the entire PDN with just one-time execution, while many other frameworks demand repeated calculations. 
\end{itemize}
Experimental results on four different commercial PDNs show that the proposed framework consistently outperforms the commercial simulator and another state-of-the-art CNN based prediction model, PowerNet~\cite{2020PowerNet}. The framework can achieve 0.63-1.02\% mean relative error with less than 1mV mean absolute error, while the framework can achieve 25-69$\times$ speed up over the commercial tool. The detailed analysis on D4 clearly shows that the predicted noise distribution is very consistent with the ground-truth, which can identify almost all the hotspots.

\section{Background}
\label{sec:background}

The core problem of PDN noise sign-off is to solve a sparse linear system after discretization, where interconnect parasitics and decoupling capacitors (decaps) constitute a symmetric and positive definite sparse matrix representing the entire power delivery system~\cite{2000HA,2003AAMA,2006RandomWalks, 2008AlgebraicMultigrid, ZhaoICCAD2014}. The instance switching draws currents from the power delivery system and is typically modelled as current sources, causing voltage droops in PDN. The PDN noise sign-off can be further categorized to static and dynamic analyses~\cite{2000HA,2003AAMA,2006RandomWalks, 2008AlgebraicMultigrid, ZhaoICCAD2014}. Static analysis employs DC excitation and hence ignores the impact of capacitance or inductance in the extracted parasitics, which is basically to solve a series of linear equations. Dynamic analysis is fed with dynamically changing current sources and models the impact of decap or even inductance. In the commercial PDN noise sign-off tools, the dynamic analysis is converted to a series of static analyses, where the system matrix is the same but with different right-hand-side items. 

For either static or dynamic analysis, the computational cost exponentially increases with the number of unknowns in the power delivery system~\cite{2003AAMA}. When dealing with tens of test vectors for dynamic analysis, it is very time-consuming to complete all the vectors by deploying the conventional simulation based methods~\cite{2000HA,2003AAMA,2006RandomWalks, 2008AlgebraicMultigrid, ZhaoICCAD2014}. Recently, various machine learning based algorithms have been proposed for static IR drop analysis~\cite{2019IncPIRD, 2020XGBIR, 2021Thermal}. For example, the work in~\cite{2019IncPIRD} proposed an incremental IR drop prediction model, which feeds the structural and electrical features of PDN into XGBoost to estimate the static IR drop. In \cite{2020XGBIR}, more features are included for static IR drop estimation to achieve higher performance. However, the overhead to obtain cell-level features for training is not well discussed in both papers, which is non-trivial in practical deployment. While the two methods focus on voltage drop estimation for each node in the power grid, a fully convolutional network based model is used to generate the entire IR drop map for PDN~\cite{2021Thermal}, whose scalability is concerning. Since the method directly converts the current/voltage maps to images, the input dimension and neural network size then restrict its application to only small PDNs. On the other hand, machine learning based dynamic analysis is also investigated in a few prior work~\cite{2018ECO, 2018MLECO, 2020PowerNet} using different machine learning models, from artificial neural network, XGBoost, to CNN. For example, \cite{2020PowerNet} utilizes power, activity, toggling rate, and neighboring information, to predict the dynamic noise. However, to limit the underlying neural network size, it has to partition the entire network to tiles and compute the noise from tile to tile for each time stamp, which is then very time consuming.

\section{proposed method}
\label{sec:design}
\subsection{Overall Flow}
\begin{figure}[h]
  \centering
  \includegraphics[width=0.85\linewidth]{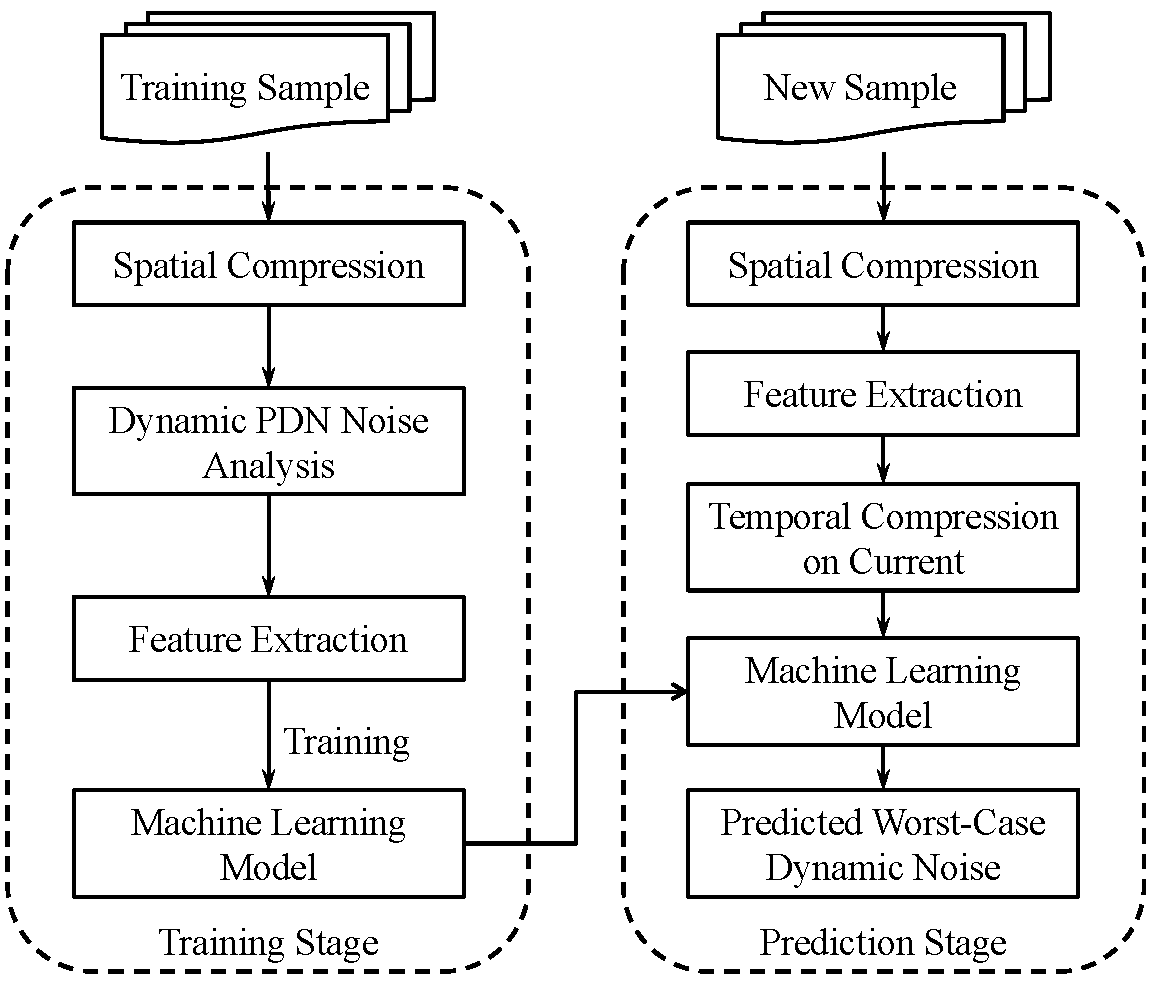}\vspace{-0.3cm}
  \caption{Overall flow of proposed framework.}\vspace{-0.3cm}
  \label{fig:flow}
\end{figure}
When deploying CNN for dynamic PDN noise prediction, it is natural to consider using the instance current map as the input and output the voltage map for the nodes in PDN. However, the huge dimensions of both input and output make such network infeasible to commercial PDNs with millions of nodes. It is noted that WNV is to ensure the worst-case noise (both spatially and temporally) to meet the pre-defined specification, i.e., 
\begin{equation}
    \max_{i\in \mathcal{N}} \max_{t\in \mathcal{S}} v_i(t) \le v_{spec}
\end{equation} 
where $\mathcal{N}$ is the set of all the nodes in a PDN and $\mathcal{S}$ refers to the time-span of the simulated trace. Obviously, WNV targets at the noise of a particular node at a particular time. It is then unnecessary to explicitly compute the noises of the rest of the nodes at every time stamp, many of which are far lower than the worst-case noise, but demand significant computational time. Thus, it is desired to filter out the unnecessary computations both spatially and temporally. 

Based on such intuitions, we designed a worst-case dynamic PDN noise prediction framework as shown in Fig.~\ref{fig:flow}. The training procedure employs a commercial PDN sign-off tool to obtain the ground-truth and train the model. The inference flow first employs a spatial and temporal compression step to remove redundancy in both input test vector (current in this paper) and PDN. Then, a feature extraction step is included to fuse different features. Instead of using instance-level fine-grained feature that needs additional extraction, we just use the same current vector information as the input to the commercial tool and physical information. Such feature selection is found to greatly reduce the training overhead. Then a CNN structure is designed to conduct the worst-case noise prediction.


\subsection{Spatial and Temporal Compression}

Instead of computing the worst-case noise for every node, we would like to merge a few nodes into a tile $T_j$ and then predict the worst case noise of the tile, where $T_j\in \mathcal{T}$ and $\mathcal{T}$ is the set of all the tiles. In other words, we can always partition the PDN layout into an array of $ m \times n $ tiles and have:
\begin{equation}
    \max_{i\in \mathcal{N}} \max_{t\in \mathcal{S}} v_i(t)= 
    \max_{T_j\in \mathcal{T}} \max_{t\in \mathcal{S}} [\max_{i\in T_j} v_i(t)]
\end{equation}
Instead of predicting $v(t)$, we can just compute $\max_{i\in T_j} v_i(t)$, which is especially important for larger PDN. Then we reduce the input and output dimensions from millions to $m\times n$.

For test vector, the sampled current map across the entire PDN can vary significantly over time. It is noted that steady state (with steady current) typically does not contribute to the worst-case noise. When the instance is heavily switching, it is more possible to inject the worst-case noise. Thus, it is necessary to filter out unimportant segments to accelerate the inference. Algorithm 1 describes how to conduct temporal feature compression in our framework, which basically acts as a classifier that retains the segments with large current or switching activity. Thus, the basic idea of the proposed algorithm for temporal compression is to remove the segments with moderate currents. According to the algorithm, we first calculate the total current at each time stamp $t_k$ and obtain the sequence $ \{S[k] \in \mathbb{R}^{m \times n} | k \in [1,N]\} $, where $N$ is the total number of time stamps. Then, we sort $\{S[k]\}$ in an ascending order and compress the segments with moderate currents to have the compressed set $\{S^{\prime}[k]\}$ with similar $ \mu+3\sigma $ to the original set. Then, we can have the compressed current maps $C_I=\{ I^{\prime}[j]\in \mathbb{R}^{m \times n} |j\in [1,r*N] \}$, where $r$ is the compression rate. The efficiency of our prediction can be improved by such spatial and temporal compression without sacrificing much accuracy. 

\begin{algorithm}[t]
\caption{Temporal Compression on Current Vector}
\flushleft{\bf Input:}
Current feature maps $ \{I[k] \in \mathbb{R}^{m \times n} | k \in [1,N]\} $, compression rate $r\in (0,1)$, rate step $\Delta r>0$ \\
\flushleft{\bf Output:}
Compressed current maps $C_I$\\
\begin{algorithmic}[1]
\State \textbf{Initialize} $d_{min}=\infty$, $r_0=0$, $C_I=\emptyset$
\For{each $I[k]$}
\State $S[k]=\sum_{x=1}^m \sum_{y=1}^n I[k][x][y]$
\EndFor
\State $\mu_s=\frac{1}{N}\sum_{k=1}^N S[k]$
\State $\sigma_s=\sqrt{\frac{\sum_{k=1}^N (S[k]-\mu_s)^2}{N}}$
\State $\{A[i]| i \in [1,N]\}$= argsort($\{S[k]| k \in [1,N]\}$)  // Sort $\{S[k]\}$ in ascending order and return the index $\{A[i]\}$
\While{$r_0\leqslant r$}
\State $C=\emptyset$
\For{each $p \in [1,r_0 * N]\cup [(1-r+r_0) * N,N]$}
\State $C=C\cup \{S[A[p]]\}$
\EndFor
\State $\mu_c=\frac{1}{r* N}\sum_{\forall S[k] \in C} S[k]$
\State $\sigma_c=\sqrt{\frac{\sum_{\forall S[k] \in C} (S[k]-\mu_c)^2}{r*N}}$
\If{$|(\mu_s+3\sigma_s)-(\mu_c+3\sigma_c)|<d_{min}$}
\State $d_{min}=|(\mu_s+3\sigma_s)-(\mu_c+3\sigma_c)|$
\State $r_s=r_0$
\EndIf
\State $r_0=r_0+\Delta r$
\EndWhile
\For{each $q\in [1,r_s*N]\cup [(1-r+r_s)*N,N]$}
\State $C_I=C_I\cup \{I[A[q]]\}$
\EndFor
\end{algorithmic}
\end{algorithm}

\begin{figure*}[t]
  \centering
  {\includegraphics[width=\textwidth]{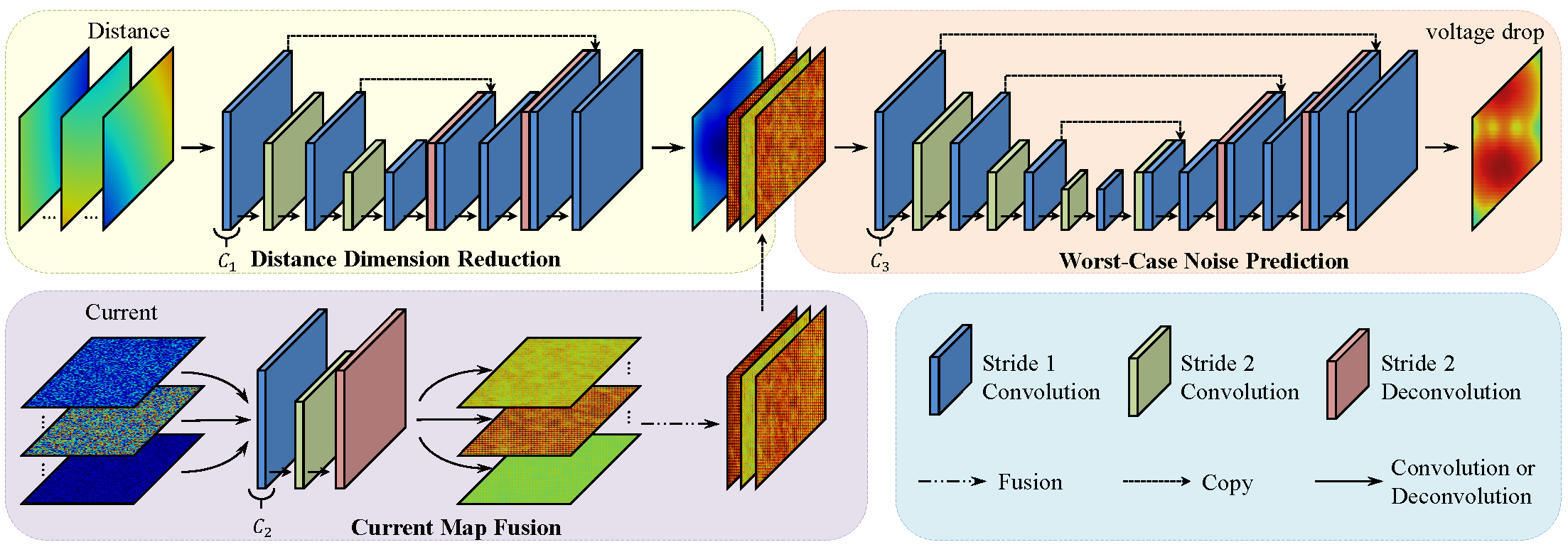}}\vspace{-0.3cm}
  \caption{The architecture of the proposed worst-case dynamic PDN noise prediction.}\vspace{-0.3cm}
  \label{fig:model}
\end{figure*}

\subsection{Feature Extraction}

While detailed features of PDN may help improve model accuracy, too many features may also lead to information redundancy and model overfitting. Appropriate feature selection is not only important to the model inference efficiency but also training overhead. To ensure the machine learning model to learn effective knowledge from input features without much training overhead, we would like to use the expert knowledge of PDN noise analysis to facilitate the feature selection:
\begin{itemize}
\item \textbf{Load current}: Similar as the commercial tool that takes current activity as the input, we choose the load current as one of the features to model the excitation to the system. The load current is organized as a feature map so as to keep the correlation among neighboring instances. When spatially compressing the PDN layout, the instance currents within a tile is summed up to compute the load current. 
\item \textbf{Distance to power bumps}: If an instance is far from the power source, the current would go through a long path and induce a larger voltage droop. While such embedded information can be extracted by learning, we choose to explicitly take the information as input to simplify the following network size. We choose the center point of a tile as representation and then compute the Euclidean distance between the center point and all the power bumps. For a PDN with $ B $ power bumps, we are supposed to calculate the distance between each pair of tile and power bump, and then assemble a distance feature matrix $ D \in \mathbb{R}^{B \times m \times n} $.
\end{itemize}
Unlike many prior works with instance-level features~\cite{2018ECO, 2018MLECO}, which demand additional simulations, the proposed feature extraction employs easily accessible features and is consistent with the conventional PDN sign-off flow. Thus, the proposed feature extraction can potentially help reduce training overhead as well as the following CNN architecture.

\subsection{Worst-Case Dynamic PDN Noise Prediction}

The architecture of the proposed worst-case dynamic PDN noise prediction model is shown in Fig.~\ref{fig:model}, which consists of three subnets to implement the distance feature processing, current map fusion, and voltage drop prediction. First, the distance feature matrix is fed into the first subnet to reduce the distance dimension. Then the current feature map is handled by a current map fusion subnet separately to obtain the fused map. Finally, the low-dimension distance map along with the fused current map is sent to the noise prediction subnet to estimate the worst-case dynamic noise.

\subsubsection{Distance Dimension Reduction}

In general, there are many power bumps in an SoC. For a particular tile, due to the locality~\cite{2010MissingLink}, only a few of the power bumps have significant impact on its worst-case noise. Here we use a U-Net like structure to achieve such reduction. The input distance map is first downsampled by convolutional layers and then upsampled by deconvolutional layers. Each of these two layers has the stride of 2 and is followed by a convolutional layer with the stride of 1. Moreover, skip connection is applied between the downsampling and upsampling features with the same size. Replication padding is applied in the convolutional layers and zero padding for the deconvolutional layers. Except the output layer, all the other layers adopt a ReLU activation function. The output layer has only one kernel so as to reduce the distance feature to $ \tilde{D} \in \mathbb{R}^{m \times n} $.

\subsubsection{Current Map Fusion}

To learn the timing information in the compressed current vector, we design a current map fusion subnet. Each sampled current map is separately sent to the network, which can handle the vector with various lengths. An encoder-decoder structure is applied to the current map fusion subnet. As the input feature has only one channel, a small network with four layers is enough. For each tile, the subnet extracts three features, the maximum value of the peak current $ \tilde{I}_{max} \in \mathbb{R}^{m \times n} $, the mean of maximum and minimum currents $ \tilde{I}_{mean} \in \mathbb{R}^{m \times n} $, and $ \mu + 3 \sigma $ as the last feature $ \tilde{I}_{msd} \in \mathbb{R}^{m \times n} $, where $\mu $ and $\sigma$ refer to mean and standard deviation of the currents.

\subsubsection{Dynamic PDN Noise Prediction}

After distance dimension reduction and current map fusion, the $ \tilde{D} $ is concatenated with $\tilde{I}_{max} $, $\tilde{I}_{mean} $ and $\tilde{I}_{msd} $ into a matrix with the size of $4 \times m \times n $. Then it is sent to the worst-case noise prediction subnet. The dynamic PDN noise prediction subnet also has a U-Net like structure.  The final output is the predicted PDN noise map $ V \in \mathbb{R}^{m \times n} $.

\subsubsection{Model Training}

To reduce the redundancy of the training set and improve efficiency of training, a training set expansion strategy is applied. If the distance between each existing sample and a candidate sample is larger than a pre-defined threshold, the candidate sample will be added to the training set. By controlling the threshold, the training set approximately accounts for 60\% in the dataset. The rest samples are randomly split into validation and test set in a ratio of 3:7. We use Adam optimizer~\cite{2014Adam} to train the proposed model with learning rate of 0.0001. The L1 loss function is adopted during training and formulated as:
\begin{equation}
    \mathcal{L}=\sum_{i=1}^{m \times n}| v_i-\hat{v}_i| 
\end{equation}

\section{Experimental Results}
\label{sec:evaluation}

\subsection{Experimental Setup}

\begin{table}[h]
\caption{Characteristics of designs in experiment.}\vspace{-0.3cm}
\label{tab:designs}
\begin{tabular}{cccccc}
\toprule
\tabincell{c}{Design}& \tabincell{c}{\#Node \\(M)}& \tabincell{c}{\#$I_{load}$\\ (k)}& \tabincell{c}{Mean WN \\(mV)}& \tabincell{c}{Max WN \\(mV)}& \tabincell{c}{Hotspot \\ratio}\\
\midrule
D1 & 0.58 & 2.5 & 100.4 & 131.7 & 56.3\% \\
D2 & 0.58 & 16.9 & 91.7 & 128.4 & 30.1\% \\
D3 & 2.67 & 122.5 & 127.1 & 290.7 & 57.5\% \\
D4 & 4.40 & 810 & 89.0 & 119.9 & 22.5\% \\
\bottomrule
\end{tabular}
\vspace{-0.3cm}
\end{table}

In our experiment, we employ four commercial PDN designs with different sizes, whose characteristics are summarized in Table \ref{tab:designs} (denoted as D1 to D4), where \#Node is the total number of the power grid and \#$I_{load}$ is the number of current loads (both are reported by an state-of-the-art commercial tool). The reported Mean and Max worst-case noise (denoted as WN in the table) across all the tiles show that the four designs have dramatically different tolerance to noise, with maximum WN varying from 12\% to 29\%. Hotspot is the number of tiles whose worst-case noise exceeds the pre-defined 10\% threshold of the nominal supply voltage (=1V). For each design, we randomly generate 500 groups of test vectors and run dynamic PDN noise simulation with the commercial tool for comparison. In the following experiments, we set the time step $\Delta t = 1$ps and the number of kernels $C_1=C_2=8$ and $C_3=16$, where $C_1$, $C_2$, and $C_3$ are the number of kernels of all layers (excluding the output layer) in distance dimension reduction, current map fusion and worst-case noise prediction subnets, respectively. The proposed framework was implemented in Python using PyTorch.

\begin{figure}[h]
  \centering\vspace{-0.15cm}
  \includegraphics[width=\linewidth]{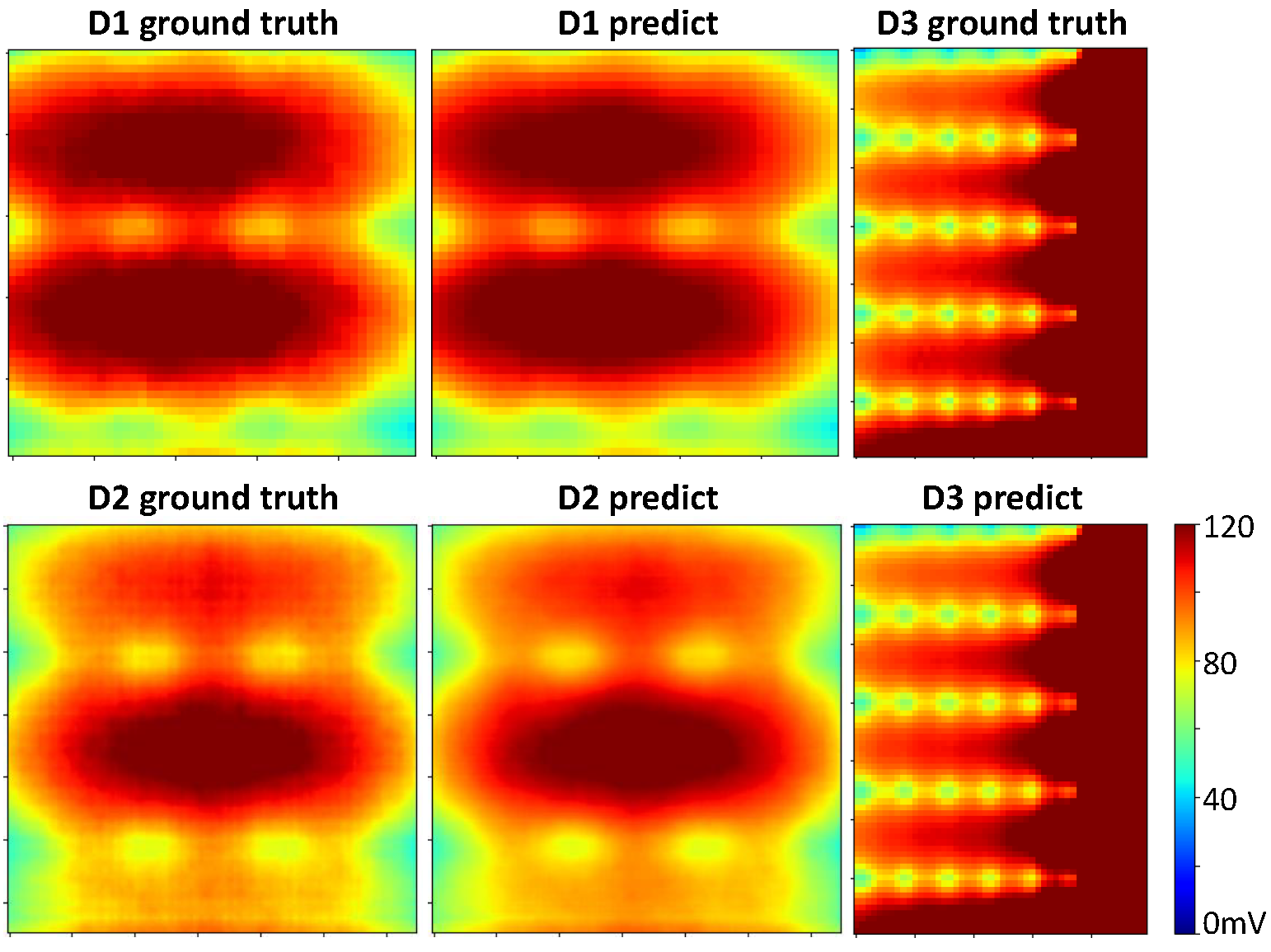}\vspace{-0.3cm}
  \caption{Comparison between the ground-truth and predicted worst-case dynamic PDN noise map for D1-D3.}
  \label{fig:noise}\vspace{-0.5cm}
\end{figure}

\begin{figure}[t]
  \centering
  \includegraphics[width=\linewidth]{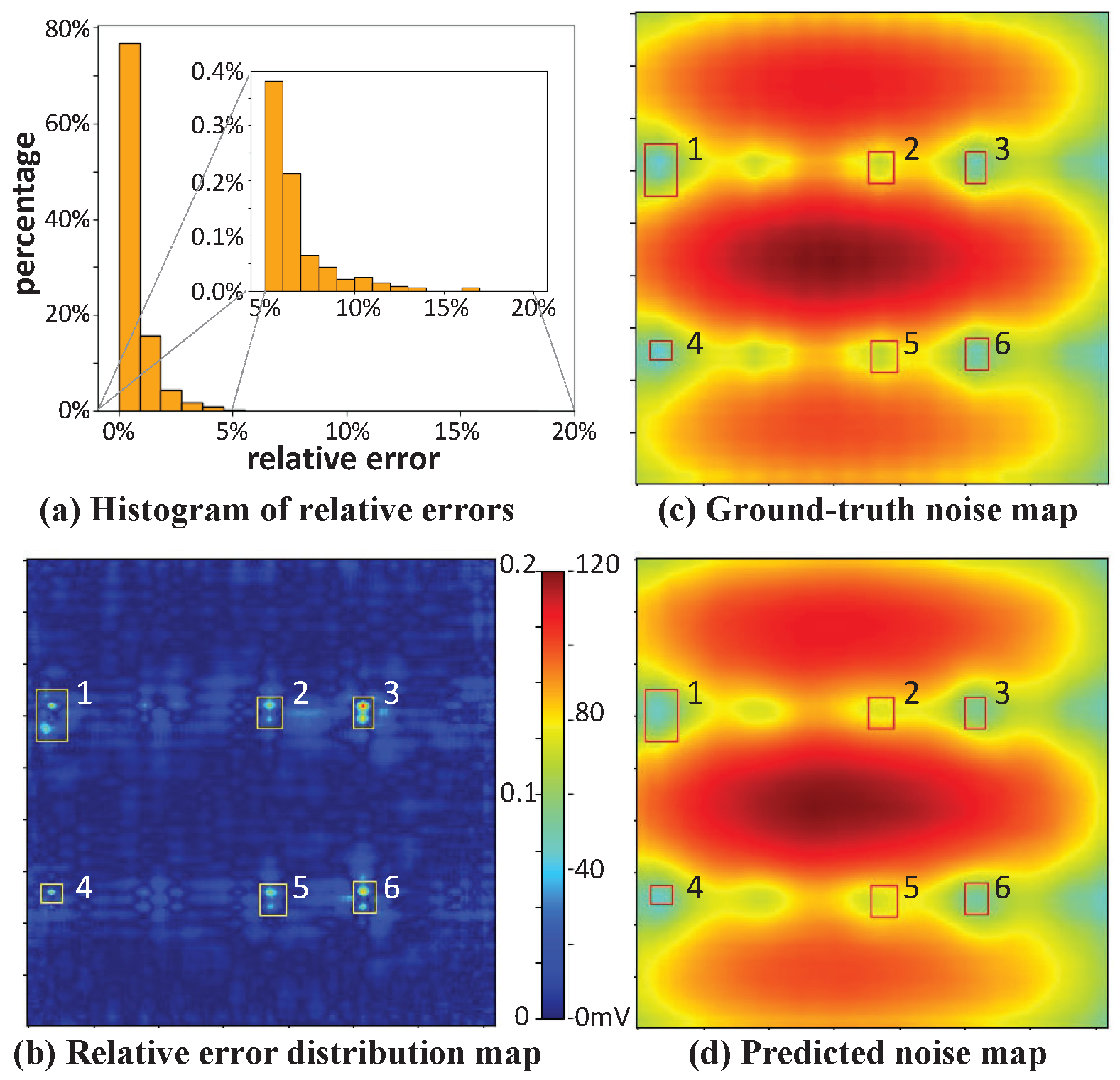}\vspace{-0.2cm}
  \caption{Prediction results of D4: (a) Histogram of relative errors; (b) Relative error distribution map; (c) Ground-truth noise map; and (d) Predicted noise map.}\vspace{-0.65cm}
  \label{fig:distribution}
\end{figure}

\begin{table*}[bp]
\caption{Comparison of accuracy and run-time between proposed framework and the commercial tool.}\vspace{-0.3cm}
\label{tab:performance}
\begin{tabular}{c|c|c|c|c|c|c|c|c}
\toprule
\multirow{2}{*}{Design}&\multirow{2}{*}{$m\times n$}&\multicolumn{3}{c|}{Accuracy evaluation}&\multicolumn{3}{c|}{Runtime comparison}&\multicolumn{1}{c}{Hotspot} \\
\cline{3-9}
& &Mean AE/RE& 99\% AE/RE&Max AE/RE&Proposed (s) &Commercial (s)&Speedup&Missing rate \\
\hline
D1&$50\times 50$&0.98mV/1.02\%&3.21mV/3.66\% &4.45mV/5.88\%& 2.93&76&26$\times$&1.09\%\\
D2&$130\times 130$&0.74mV/0.83\%&2.35mV/2.89\% &4.12mV/5.83\%&2.72&69&25$\times$&1.95\%\\
D3&$70\times 50$&0.71mV/0.63\%&3.00mV/3.47\% &9.38mV/10.53\%&4.60&320&69$\times$&0.28\%\\
D4&$180\times 180$&0.58mV/0.71\%&2.90mV/4.20\% &8.23mV/16.80\%&8.95&621&69$\times$&1.93\%\\
\bottomrule
\end{tabular}
\end{table*}

\subsection{Framework Evaluation}

Table~\ref{tab:performance} summarizes the prediction result of the proposed framework in comparison to the commercial tool. The accuracy of the proposed framework are evaluated with mean absolute error (denoted as AE), mean relative error (denoted as RE), 99\% percentile AE and RE, and maximum AE and RE. As shown in the table, the proposed framework can achieve a mean AE less than 1mV and 0.63-1.02\% mean RE. Moreover, even for 99\% percentile, AEs are only 2-3mV with REs of 2.9-4.2\%. Though D4 reports larger maximum RE of 16.8\%, the corresponding AE is only around 8mV, whose tile has a small worst-case noise. Thanks to the accuracy of the proposed framework, almost all the hotspots for the four designs are correctly identified with merely 0.28-1.95\% missing rate, as shown in the last column of the table. On the other hand, the run-time of the proposed framework is much faster than the commercial tool, with 25-69$\times$ speed up. 

The predicted worst-case PDN noise distribution maps and their ground-truths for D1, D2 and D3 are shown in Fig.~\ref{fig:noise}. The predicted results are almost identical to the ground-truths, which indicates that the proposed framework can effectively replace the commercial tool to support WNV for different vectors. The detailed prediction results of D4 are demonstrated in Fig.~\ref{fig:distribution}. Fig.~\ref{fig:distribution}(a) shows the histogram of the relative errors for all the tiles in D4. It is clear that most of the tiles have relative errors of less than 5\%. Only very few tiles have small worst-case noise and hence result in larger relative errors. Fig.~\ref{fig:distribution}(b) shows the noise map of the relative error, where the marked spots are the tiles with larger relative errors. Fig.~\ref{fig:distribution}(c) and \ref{fig:distribution}(d) present the comparison between the predicted and simulated voltage maps. Similar as Fig.~\ref{fig:noise}, the two subfigures are almost identical. It is also observed that the corresponding marked spots have low noise, which are then not the concerned spots in PDN designs.


We also compare the proposed framework with PowerNet~\cite{2020PowerNet} on D4, which is a CNN based tool to predict dynamic PDN noise. The internal and leakage power, signal arrival time, and toggling rate are extracted and fed to PowerNet~\cite{2020PowerNet}. We set the number of time-decomposed power maps as 40, input window size as 15, and then partition the PDN design similarly ($180 \times 180$ tiles) as the proposed framework. PowerNet is trained with the same data as the proposed framework for fair comparison. Table \ref{tab:powernet} shows that the proposed framework outperforms PowerNet in both accuracy and run-time, with almost 20$\times$ improvement in mean AE and 3$\times$ speed up. By comparing AUC between the two, the proposed framework can achieve almost 40\% AUC improvement.

\subsection{Temporal Compression}

\begin{figure}[h]
  \centering\vspace{-0.2cm}
  \includegraphics[width=\linewidth]{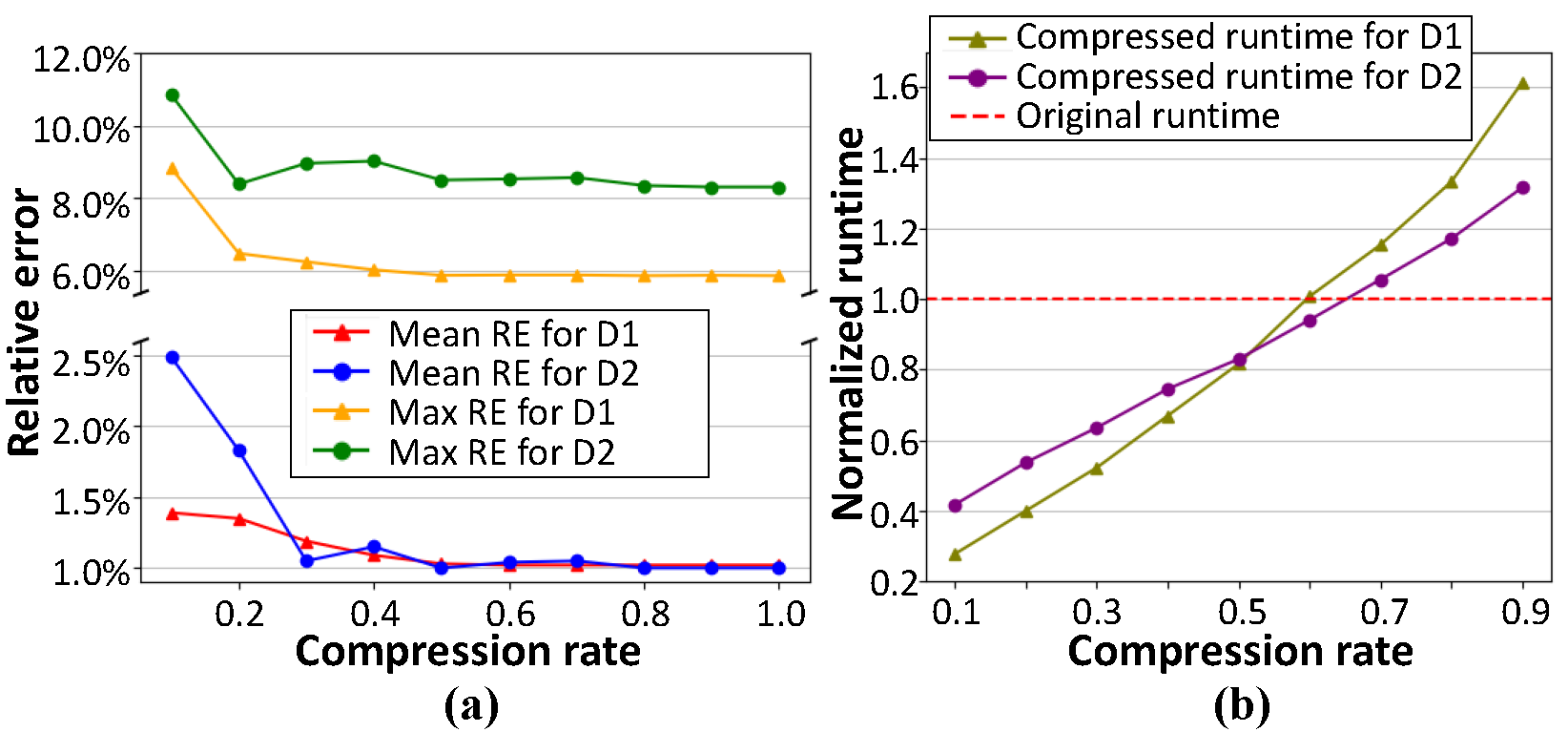}\vspace{-0.3cm}
  \caption{Impact of the proposed temporal compression algorithm: (a) Relative error change $w.r.t.$ compression rate; and (b) Run-time $v.s.$ compression rate.}\vspace{-0.5cm}
  \label{fig:compress}
\end{figure}

Fig.~\ref{fig:compress} summarizes the performance of the proposed temporal compression algorithm. Fig.~\ref{fig:compress}(a) shows how the relative error changes with different compression rate. Though the errors generally drops with a larger compression rate, i.e., more data are retained, there is a knee point after which the degradation is much faster. For D1 and D2, such a compression rate threshold is around 0.3, with mean relative errors of 1.19\% and 1.05\% for D1 and D2, respectively. The run-time comparison after using the proposed temporal compression is shown in Fig.~\ref{fig:compress}(b). With a higher compression rate, the run-time is almost linearly increased.

\begin{table}[t]
\caption{Comparison between the proposed framework and PowerNet~\cite{2020PowerNet}.}\vspace{-0.3cm}
\label{tab:powernet}
\begin{tabular}{cccccc}
\toprule
\tabincell{c}{Model}& \tabincell{c}{MAE\\ (mV)}& \tabincell{c}{Mean\\ RE}& \tabincell{c}{Max\\ RE}& \tabincell{c}{AUC} & \tabincell{c}{runtime \\(s)} \\
\midrule
PowerNet~\cite{2020PowerNet} & 11.69 & 13.71\% & 42.08\% & 0.602 &23.25\\
Ours & 0.58 & 0.71\% & 16.80\% &0.999& 8.95\\
\bottomrule
\end{tabular}
\vspace{-0.4cm}
\end{table}

\section{Conclusions}
\label{sec:conclusions}

This paper presents a worst-case dynamic PDN noise prediction framework using a CNN. The framework employs spatial and temporal compression to reduce the input and network complexity. An expert knowledge guided feature extraction is then used to further simplify the network structure. Finally, a prediction model is designed with three subnets to predict the worst-case PDN noise. Experimental results show that the proposed framework beats both the commercial tool and a recently reported solver with 1-2 orders of magnitude improvements, which has good scalability and efficiency while maintaining high accuracy.

\begin{acks}
We would like to thank the insightful discussions with Dr. Jun Chen, Dr. Yucheng Wang and Dr. Ji Li from Giga Design Automation Co., Ltd. This work was supported in part by Zhejiang Provincial Key R\&D program (Grant No. 2020C01052), NSFC (Grant No. 62034007 and 62141404) and Guangdong Provincial Key R\&D program (Grant No. 2021B1101270003).
\end{acks}

\end{document}